\documentclass[journal]{IEEEtran}
\usepackage[ruled]{algorithm2e}
\usepackage{booktabs}


\usepackage{subcaption}

\usepackage{cite}
\usepackage{url}
\usepackage{hyperref} 
\usepackage{ulem} 

\ifCLASSINFOpdf
  \usepackage[pdftex]{graphicx}
\else
\fi

\usepackage{amsmath}
\usepackage{xcolor}

\newcommand{\LUGLLightGBM}{LUGL-Q-LightGBM}
\newcommand{\LUGLDLightGBM}{LUGL-QD-LightGBM}
\newcommand{\LUGLPILightGBM}{LUGL-PI-LightGBM}
\newcommand{\LUGLVLightGBM}{LUGL-V-LightGBM}
\newcommand{\LUGLDeepCFRLightGBM}{LUGL-DeepCFR-LightGBM}
\newcommand{\LUGLDeepCFRMultiSD}{LUGL-DeepCFR-Multi-S/D}
\newcommand{\LUGLDeepCFRMultiS}{LUGL-DeepCFR-Multi-S}
\newcommand{\LUGLDeepCFRMultiD}{LUGL-DeepCFR-Multi-D}
\newcommand{\oneStepLookaehadFunction}{T}

\usepackage{amsfonts}
\usepackage{tikz}
\usetikzlibrary{positioning, arrows, matrix}

\tikzstyle{box} = [draw, rounded corners, minimum height=1cm, minimum width=2.5cm, align=center]
\tikzstyle{arrow} = [->, thick]
\tikzstyle{dasharrow} = [->, thick, dashed]
\begin{document}
%
\title{Local Updates, Global Learning (LUGL): Playing Games with non-incremental Learners }
%
%

\author{ David Milec, Spyridon Samothrakis, Michael Fairbank, and Dennis J.N.J. Soemers%

\thanks{David Milec is with the Faculty of Electrical Engineering, Czech Technical University in Prague, Prague, Czech Republic (e-mail: milecdav@fel.cvut.cz).} \thanks{Spyridon Samothrakis and Michael Fairbank are with the School of Computer Science and Electronic Engineering, University of Essex, Colchester, United Kingdom (e-mail: ssamot@essex.ac.uk; m.fairbank@essex.ac.uk).} \thanks{Dennis J.N.J. Soemers is with Maastricht University, the Netherlands (e-mail: dennis.soemers@maastrichtuniversity.nl) } \thanks{Relevant github repos are: 
\url{https://github.com/milecdav/Deep-CFR-2025}, \url{https://github.com/milecdav/PokerRL-2025}, 
\url{https://github.com/ssamot/luglrl}, \url{https://github.com/ssamot/fil}}}


\maketitle

\begin{abstract}
The dominance of Neural Networks (NNs) in RL is partially due to their incremental learning capability, which naturally suits the online, non-stationary nature of self-play training. However, gradient-boosted trees like LightGBM are widely recognised as the state of the art for tabular data in supervised learning, often outperforming NNs in accuracy and efficiency. Game states are inherently tabular---discrete actions, categorical card identities, structured board positions---which makes them an ideal candidate for tree-based methods. We introduce LUGL (Local Updates, Global Learning), a framework that decouples data collection from model fitting, enabling non-incremental learners such as GBTs to operate in RL settings where they would otherwise fail due to distributional shift. LUGL alternates between a \emph{local updates} phase, where the agent plays self-play games and accumulates tabular updates (Q-values, V-values, policies, or regret values) in a finite table, and a \emph{global learning} phase, where the table is used to train a function approximator that generalises to unseen states before the table is reset. We test our approach in four standard perfect-information games (Tic-tac-toe, Connect-4, Othello, and Hex) and five imperfect-information games (Kuhn's poker, Leduc Hold'em, Liar's Dice, Goofspiel, and Flop5 Hold'em), and show that our results are competitive with or superior to DQN and DeepCFR. Our experiments demonstrate that the community's strong bias towards NNs in game-playing may be unwarranted, since LightGBM-based agents achieve competitive or superior performance across all tested benchmarks. 
\end{abstract}


%
\IEEEpeerreviewmaketitle

\section{Introduction}

Game playing has been a popular research area in the field of artificial intelligence (AI) for decades. There are multiple reasons for this, but primarily, games offer a clean interface where one can test Reinforcement Learning (RL) and tree-search algorithms in popular domains, without having to deal with issues that stem from not having a perfect model to simulate future trajectories. That is, one can be sure that doing well in the simulation entails doing well in the problem that is currently studied. Games also tend to expose an abstraction of state (i.e., how the world looks) and allow an agent to traverse it, often with surprising computational efficiency.


Although there is a whole zoo of gaming-playing algorithms, two of the more impactful ones are arguably variations of deep Q-Learning (DQN)~\cite{Mnih_2015_DQN} and Deep Counterfactual Regret Minimization (DeepCFR)~\cite{brown2019deep}, for perfect and imperfect-information games, respectively. Though we have seen other algorithms potentially having a bigger impact (e.g. RL had significant successes with actor-critic/policy gradient like methods like PPO~\cite{schulman2017proximal}), these two variations also seem to be some of the most fundamental and easy to implement. We have also had further variations of both DQN and CFR that have created significant breakthroughs in game-playing algorithms~\cite{anthony2017thinking,silver2017mastering,silver2018general, browne2012survey}. The substrate of these game-playing algorithms is a combination of tree searches (with all kinds of variations coming in, e.g. search in abstracted spaces, search using Monte-Carlo methods) and a function approximator that tries to approximate some form of value function, i.e., how good an information set (or state) is for an agent, without conducting a full evaluation.

The main advantage of using NNs over any other function-approximation methods is their ability to learn incrementally. That is, one can train a network a bit given some data, play the game some more, collect data, and use it to train the network a bit more, repeating until convergence. Since NNs are not immune at all to distributional shifts, in DQN the training data are kept in a buffer called ``Experience Replay''\cite{DBLP:journals/corr/SchaulQAS15}, one network is playing, while a background network is training; and at some point they are swapped. Incremental learning, however, is not universally needed, and one might argue that the replay buffer is a hack to help the NN maintain a semblance of stability. In DeepCFR, one re-learns the network from scratch following a data collection process, which is itself closer in spirit to a batch learning approach.

Despite the incremental learning advantage of NNs, tree-based methods such as Gradient Boosting~\cite{ke2017lightgbm} are widely recognised as the state of the art for tabular data in supervised learning, often outperforming NNs in accuracy and efficiency~\cite{NEURIPS2022_0378c769}. Game states are inherently tabular---discrete actions, categorical card identities, structured board positions---which makes them a natural candidate for GBTs. They are almost non-parametric, have fewer hyperparameters to tune, and provide stronger inductive bias and built-in variance control. This ``ease of use'' makes them the method of choice for supervised-learning tasks, and they are widely used in industry. The problem we aim to solve is therefore twofold: (i) provide a principled framework (LUGL) that enables batch learners to operate in RL settings where they would otherwise fail due to the non-stationary, evolving distribution of states and targets generated by self-play; and (ii) demonstrate empirically that the community's strong bias towards NNs in game-playing may be unwarranted.

How far can we steer from the incremental learning and NNs without causing too much trouble? This paper introduces LUGL, an algorithm for training non-incremental models in game-playing settings. The LUGL algorithm is a novel approach that combines ideas from fitted Q-iteration, approximate policy iteration, DQN and DeepCFR. It involves a combination of local updates and global learning, where an agent gathers data locally during self-play and is then globally updated using the collected self-play data. The algorithm works by iteratively collecting a table of action outcomes over a number of games (this is the ``Local Updates'' step), and then periodically uses this table to train a function approximator, usually a LightGBM, to generalise into a function defined over the whole game space (e.g. a value or a policy function); this is the ``Global Learning'' step. Once the function approximator is created, the local table is reset, and the function approximator is used to drive the agent policy in more game play; and the whole process iterates: building up the local table; training a function approximator from that table; resetting the local table.

This paper tests the algorithm on four perfect-information games, namely Tic-tac-toe, Connect-4, Othello, and Hex, and five imperfect-information games, Kuhn's and Leduc's poker, Liar's Dice, Goofspiel and Flop5 Hold’em. The performance of LUGL is compared with the DQN and DeepCFR algorithms. We explore a number of LUGL variations, showing that all are reasonably robust. While we use some of the more fundamental algorithms to compare and build up our agents, our methods should transfer to more advanced setups.

Our main contributions are as follows: (a) We introduce LUGL, a framework that decouples data collection from model fitting, enabling any batch learner---such as gradient-boosted trees, splines, or decision trees---to replace neural networks in game-playing RL. (b) We provide the first controlled comparison of function approximators within DeepCFR under identical training budgets, showing that LightGBM consistently outperforms neural networks across five benchmark games. (c) We demonstrate that neural networks are not necessary for competitive play in either the DQN or DeepCFR paradigm: LUGL variants converge faster than DQN in perfect-information games and achieve lower exploitability than DeepCFR in all tested imperfect-information games, including a ${\sim}100$\,mbb/h (milli-big-blinds per hand). (d) We provide evidence that the non-stationary, noisy regression targets inherent to CFR-based training favour tree-based models, whose stronger inductive bias and built-in variance control yield more stable learning than gradient-descent-trained networks.
Although our experiments employ well-established baselines (DQN, DeepCFR, SD-CFR), the LUGL framework is algorithm-agnostic: any RL method whose training decomposes into a data-collection phase and a supervised-learning phase is a candidate for the non-incremental treatment we propose.

The paper is structured as follows. Section II provides a brief overview of game-playing algorithms and their use of NNs, also discussing the limitations of NNs in game-playing algorithms and the potential advantages of non-neural methods such as GBTs. Section III introduces the LUGL algorithm and its design principles. Section IV presents the experimental setup, including the games used for testing and the comparison with DQN, and a comparison of LUGL variants with standard NN algorithms. Finally, Section V concludes the paper and discusses the implications of the results.

\section{Background} \label{Sec:Background}

In this section, we briefly discuss the RL concepts used throughout the paper, as well as some of the sample games.

\subsection{Game abstractions}

\subsubsection{Two-Player Extensive-Form Games (POMDP-style notation)}

A two-player extensive-form game is given by the tuple
\[
\mathcal{G} = \big(\mathcal{H},\, P,\, \mathcal{A}_1,\, \mathcal{A}_2,\, \mathcal{I}_1,\, \mathcal{I}_2,\, u_1,\, u_2 \big),
\]
with components described as follows:
\begin{itemize}
  \item $\mathcal{H}$ is the set of histories (finite sequences of past actions and chance outcomes). Terminal histories form $\mathcal{Z}\subseteq\mathcal{H}$.
  \item $P:\mathcal{H}\setminus\mathcal{Z}\to\{1,2,c\}$ is the player function: $P(h)=i$ if player $i$ acts at history $h$, and $P(h)=c$ for a chance node.
  \item $\mathcal{A}_1$ and $\mathcal{A}_2$ are the (finite) action sets of players 1 and 2, respectively. For a history $h$ with $P(h)=i$ we denote the available actions by $\mathcal{A}_i(h)\subseteq\mathcal{A}_i$.
  \item $\mathcal{I}_1$ and $\mathcal{I}_2$ are partitions of $\{h\in\mathcal{H}:P(h)=1\}$ and $\{h\in\mathcal{H}:P(h)=2\}$ respectively; each $\mathcal{I}_i$ is the collection of information sets for player $i$. For any $I \in \mathcal{I}_i$, if $h,h'\in I$, we require $\mathcal{A}_i(h)=\mathcal{A}_i(h')$.
  \item $u_1,u_2:\mathcal{Z}\to\mathbb{R}$ are the utility (payoff) functions for players 1 and 2.
\end{itemize}

\subsubsection{Two-Player Alternating Markov Games}

A two-player alternating Markov game (AMG) is the following restricted form:
\[
\mathcal{M} = \big(\mathcal{S},\, \mathcal{A}_1,\, \mathcal{A}_2,\, T,\, R_1,\, R_2,\, \gamma,\, \rho_0\big),
\]
in which:
\begin{itemize}
  \item $\mathcal{S}$ is the state space. There is a surjective map $\varphi:\mathcal{H}\to\mathcal{S}$ that relates histories to states in the induced extensive form representation (Markov reduction).
  \item $\mathcal{A}_1,\mathcal{A}_2$ are the action sets for players 1 and 2. For $s\in\mathcal{S}$ and $i\in\{1,2\}$ we may write $\mathcal{A}_i(s)\subseteq\mathcal{A}_i$ for the actions available to player $i$ when the current state is $s$ (equivalently for any $h$ with $\varphi(h)=s$).
  \item $T:\mathcal{S}\times(\mathcal{A}_1\cup\mathcal{A}_2)\to\Delta(\mathcal{S}\times\{1,2\})$ is the transition function: given current state $s$ and an action $a_i\in\mathcal{A}_i$ by the active player $i$, $T(s,a_i)$ is a probability distribution over next-state/next-player pairs $(s',j)$ with $j\in\{1,2\}$ (with $\Delta$ being shorthand for the set of all probability distributions over a given set).
  \item $R_1,R_2:\mathcal{S}\times\mathcal{A}_i\to\mathbb{R}$ give the immediate rewards for the acting player (for $i\in\{1,2\}$ we write $R_i(s,a_i)$).
  \item $\gamma\in[0,1]$ is the discount factor.
  \item $\rho_0\in\Delta(\mathcal{S})$ is the initial state distribution.
\end{itemize}

An AMG is simply an extensive-form game where both players act one after another, and there is no hidden information. Below, we will see some examples of these games in the strictly competitive setting. Though AMGs are a subset of extensive-form games, it is worth breaking down game categories slightly differently, as the special properties of AMGs allow for the use of standard reinforcement learning algorithms. Thus, in what follows, we will use ``imperfect information'' to mean extensive-form games that are not trivial in that sense, and perfect-information games to mean AMGs.

\subsection{Games}
We briefly discuss the actual games we tested on.
Table~\ref{tab:ii_games_sizes_perfect} provides an indication of the complexity of the perfect-information games, in which state-space complexity refers to the number of legal positions reachable from the initial position, while the complexity of the game tree is the number of leaf nodes in the smallest full-width decision tree.

\begin{table}[h!]
\centering
\normalsize
\caption{Sizes of described perfect-information games.}
\label{tab:ii_games_sizes_perfect}
\begin{tabular}{lcc}
\toprule
\textbf{Game} & \textbf{State-Space} & \textbf{Game-Tree}  \\
 & \textbf{Complexity} & \textbf{Complexity} \\
\midrule
Tic-Tac-Toe & $10^3$ & $10^5$ \\
Connect-4 & $10^{13}$ & $10^{21}$  \\
Othello & $10^{28}$ & $10^{58}$  \\
Hex (11$\times$11) & $10^{56}$ & $10^{98}$  \\
\bottomrule
\end{tabular}
\end{table}

\subsubsection{Perfect information}

Tic-tac-toe is a simple but classic game. Players take turns placing Xs and Os on a 3$\times$3 grid, trying to get three in a row. Despite its simplicity, the game can be quite strategic, especially when played by skilled players. However, with perfect play from both sides, the game will always end in a tie, making it a solved game from an AI perspective.

Connect-4 is another popular game that is played on a larger 6$\times$7 grid, where the objective is to get four in a row. Players take turns dropping coloured discs into the columns of the grid, trying to connect four of their own colours either vertically, horizontally, or diagonally. The game is similar to tic-tac-toe in that it can also be solved using AI. With perfect play from both sides, the game will always end in a draw.

Othello, also known as Reversi, is a two-player game played on an 8$\times$8 board. Players take turns placing pieces of their colour on the board, with the goal of flipping their opponent's pieces to their own colour. The player with the most pieces of their colour at the end of the game wins. Othello has been solved for smaller board sizes using AI, with the 8$\times$8 board being solved in 2024~\cite{takizawa2023othello}. The game is complex and requires a lot of strategic thinking, making it a popular choice for competitive play.

Hex is a two-player game played on a hexagonal board, where players try to connect opposite sides of the board with their colour. The game is simple to learn, but can be difficult to master due to its many possible configurations. Hex can be proven to be a win for the first player under optimal play (or for the second player if the ``swap rule'' is used)~\cite{D-1164}, but what optimal play looks like in every state (or even just the initial state) is not yet known. Current machines are superhuman.

\subsubsection{Imperfect information}

Sizes of the imperfect-information games are given in Table~\ref{tab:ii_games_sizes_imperfect}.

\begin{table}[h!]
\centering
\normalsize
\caption{Sizes of described imperfect-information games. }
\label{tab:ii_games_sizes_imperfect}
\begin{tabular}{lcc}
\toprule
\textbf{Game} & \textbf{Total Nodes} & \textbf{Infosets per player} \\

\midrule
Kuhn Poker       & $55$    & $6$   \\
Leduc Poker      & $9,451$ & $468$ \\
Liar's Dice      & $8,177$ & $512$ \\
Goofspiel-4   & $2,229$ & $369$ \\
Flop5 Hold'em & $41 \cdot 10^{12}$ & $8 \cdot 10^9$ \\
\bottomrule
\end{tabular}
\end{table}

Kuhn Poker is the simplest variant of poker, where players are dealt one private card from a three-card deck (jack, queen, and king). The game features a single betting round with a limit of one bet of a fixed amount.

Leduc Poker extends this complexity by using a six-card deck (two jacks, two queens, and two kings), dealing each player one private card, and revealing a single public card after the first betting round. The game includes two betting rounds with limited bet sizes, making it more complex than Kuhn while remaining computationally tractable.

Both games capture the fundamental challenges of poker-balancing value betting and bluffing-despite their simplified structures. Their mathematical tractability allows for optimal solutions using computational methods, making them established benchmarks for evaluating poker AI algorithms and game-theoretic solution concepts.

Liar's Dice (Single Die, One Round) is a streamlined version of the classic bluffing game where each player receives one four-sided die that they roll and keep hidden. After examining their private die result, players take turns making bids about the total count of specific face values across all dice in the game. Each bid must either increase the quantity or the face value from the previous bid. The round continues until a player challenges the current bid by calling ``liar,'' at which point all dice are revealed to determine if the bid was accurate or not. The challenger wins if the bid was false, while the bidder wins if it was true or exceeded.

Goofspiel-4 (Imperfect Information) is a simultaneous-bid card game played with a deck of four cards (typically numbered 1 through 4) for each player, plus a separate prize deck of the same four values. At the start of each round, the top prize card is revealed, and both players simultaneously choose one card from their hand to bid for it. However, unlike the standard version, the bid cards are not revealed to the opponents - players only learn the outcome of each auction (whether they won, lost, or tied for the prize card) without seeing what their opponent actually bid. The player with the higher bid wins the prize card and adds its value to their score, while both bid cards are discarded. If players bid the same value, the prize card is typically discarded with no points awarded to either player. The game continues for four rounds until all cards are played, and the player with the highest total score is declared the winner.

Flop5 Hold'em follows the same betting rules as the full Limit Texas Hold'em, with one change. Instead of four betting rounds with some cards revealed between each, we have two betting rounds and reveal all five cards at once. The game is played with a complete 52-card deck, and each player receives two hole cards. Versions of the game are popular in the DeepCFR literature \cite{brown2019deep,steinberger2019single,steinberger2020dream}

\subsection{Comparing agents}
Evaluating agents resulting from a learning process is not trivial, and it is often different in perfect vs imperfect-information games. Though we will not go over the details as to why this is the case, we used two different evaluation systems: \emph{Glicko-2} for perfect-information games, and \emph{Exploitability} for imperfect-information games.

\subsubsection{Glicko-2}

Glicko-2 is a rating system that is designed to estimate the skill level of players in games and sports. The basic idea behind Glicko-2 is to estimate the true skill level of each player based on their observed performance in matches. This is done by assigning a rating to each player, which represents their skill level relative to the rest of the player population. The rating is represented by a number, which can be thought of as a measure of the player's ability, with higher numbers indicating higher skill.

The Glicko-2 system uses two main components to estimate player ratings: a rating and a rating deviation. The rating estimates the true skill level of each player, while the rating deviation estimates the uncertainty in that rating, based on the number of games played and the variability of the player's performance. When a player's rating is updated, the rating deviation adjusts the size of the rating change - larger deviations allow for bigger swings, while smaller deviations produce more conservative updates. The rating deviation itself is updated in two steps. Between rating periods (when a player is inactive), it inflates to reflect increased uncertainty, at a rate governed by a system constant that controls overall volatility and hence the speed at which ratings change over time. After an active period in which games are played, the rating deviation is reduced as new information is incorporated.

\subsubsection{Exploitability}

In a two-player zero-sum game, the \textit{exploitability} of a policy profile $\pi = (\pi_1, \pi_2)$ measures how much a fully rational opponent could improve their outcome by unilaterally deviating from the given policies. Intuitively, it quantifies how far a policy profile is from a Nash equilibrium: the higher the exploitability, the easier it is for an opponent to take advantage of weaknesses in the strategy.

Formally, the exploitability $\mathcal{E}(\pi)$ of a policy profile $\pi$ is defined as:
\[
    \mathcal{E}(\pi) = \max_{\pi_1^*} u_1(\pi_1^*, \pi_2) + \max_{\pi_2^*} u_2(\pi_1, \pi_2^*),
\]
where $\pi_i^*$ is the best-response policy for player $i$ against the opponent's fixed policy $\pi_{-i}$.

In our experiments, exploitability is computed directly using OpenSpiel's built-in exploitability calculation at each training checkpoint~\cite{lanctot2019openspiel}. This procedure solves for the best-response strategy against the learned policy via backward induction over the game tree and returns the resulting expected payoff gap. For the smaller games (Kuhn Poker, Leduc Poker, Liar's Dice, and Goofspiel), the state space is sufficiently small that exact best-response computation is feasible at every evaluation point. This is a pointwise measurement: a lower exploitability value at any given iteration means the policy is closer to a Nash equilibrium at that specific point in training, not merely asymptotically.

Unlike rating systems such as Glicko-2, where higher values are better, here lower exploitability is better, as it means that there is less room for an opponent to improve by deviating. An exploitability of zero corresponds to a Nash equilibrium, meaning that neither player can unilaterally improve their outcome.

\subsection{Gradient Boosting Machines}

Gradient boosting~\cite{friedman2001greedy} is a powerful machine-learning technique that belongs to the ensemble learning family; we use it as a reasonable alternative to NNs, as it is often the default algorithm used on tabular data by practising data scientists. It is particularly effective for regression and classification problems. The essence of gradient boosting lies in combining multiple weak predictive models, typically decision trees, to create a stronger and more accurate predictive model. The process starts by fitting an initial weak model, represented by a decision tree $h_0(x)$, to the data. Subsequent weak models, $h_i(x)$, are then iteratively built to correct the mistakes of the previous models.

The models are trained by minimising a predefined loss function, such as the mean squared error for regression, or the logarithmic loss for classification. The loss function is typically defined as a comparison between the predicted output $F_{m-1}(x)$ and the true output $y$. For example, the mean squared error loss can be defined as: $ L(y, F_{m-1}(x)) = \frac{1}{2}(y - F_{m-1}(x))^2 $


During each iteration, a new weak model $h_i(x)$ is trained to approximate the negative gradient of the loss function with respect to $F_{m-1}(x)$, i.e. such that $h_i(x)\approx  -\nabla L(y, F_{m-1}(x))$; and then the model is updated to $F_{m}(x)= F_{m-1}(x) + \alpha h_i(x)$, where $\alpha>0$ is a learning rate (often called shrinkage).

Decision trees are commonly used as the weak models within gradient boosting, due to their ability to capture complex non-linear relationships in the data. They can handle a mix of continuous and categorical features, and are robust to outliers. Moreover, decision trees are computationally efficient and can handle large datasets. By iteratively adding decision trees to the ensemble, gradient enhancement optimises the overall model by minimising the loss function, resulting in improved accuracy and predictive performance.

Although there are many popular implementations of gradient-boosting variants, we use LightGBM~\cite{ke2017lightgbm} in this paper. Overall, gradient boosting has been underexplored in RL, with little work done in the past (but see~\cite{abel2016exploratory} for an exception).

\subsection{DQN}

Deep Q-Networks (DQN) is a reinforcement learning algorithm that uses deep neural networks to approximate the optimal action-value function, $Q^*(s, a)$, which denotes the expectation of the total discounted cumulative reward, starting from state $s$ and taking initial action $a$ and optimal actions thereafter, and builds on top of Q-Learning~\cite{watkins1992q}. Although Q-learning was initially developed for single player environments, it can be proven to work in AMGs\cite{littman1996algorithms}. DQN combines Q-learning with deep learning, enabling it to handle high-dimensional state spaces, such as the ones encountered in perfect-information games. The algorithm uses a replay buffer to store transitions and sample mini-batches for training, which helps to break the correlation between consecutive samples and stabilises learning. Additionally, DQN employs a target network to decouple the learning of the action-value function from the evaluation of the action-value function, further enhancing stability and performance. The policy followed by DQN is typically an $\epsilon$-greedy policy, which balances exploration and exploitation by choosing a random action with probability $\epsilon$ and the greedy action with probability $1 - \epsilon$.

\subsection{Deep Counterfactual Regret Minimization (DeepCFR)}

Deep Counterfactual Regret Minimization (DeepCFR) \cite{brown2019deep} is a reinforcement learning algorithm specifically designed for games with imperfect information, such as poker. The algorithm combines the theoretical foundations of Counterfactual Regret Minimization (CFR)~\cite{zinkevich2007regret} with deep-learning techniques to approximate both cumulative regret and player policies.

\subsubsection{Core Architecture}

DeepCFR employs neural networks to represent each player's cumulative regret, from which the current policy is derived. The network parameters are updated through gradient descent optimisation throughout the learning process.

\subsubsection{Algorithm Process}

The algorithm operates through the following iterative process: (a)  Trajectory Sampling: In each iteration, DeepCFR samples $K$ trajectories using an external sampling scheme. This sampling strategy explores all possible actions for the currently active player while sampling only a single action for all other players, including chance events. (b) Policy Computation and Data Collection: For these sampled trajectories, the algorithm computes the current policy using the cumulative regret network. It then calculates immediate regret values and stores both the current policy and immediate regret data in separate reservoir buffers for later use. (c) Network Retraining: At the end of each iteration, the cumulative regret network is retrained from scratch using the immediate regret samples from the reservoir buffer \cite{brown2019deep}. (d) Final Policy Learning: Once the main training process is complete, the collected policy samples are used to train a separate policy network that represents the average policy across all iterations.

This approach allows DeepCFR to handle the complexity of large imperfect-information games by leveraging the approximation power of deep neural networks while maintaining the theoretical guarantees of CFR.


\section{Methodology} \label{Sec:Methodology}
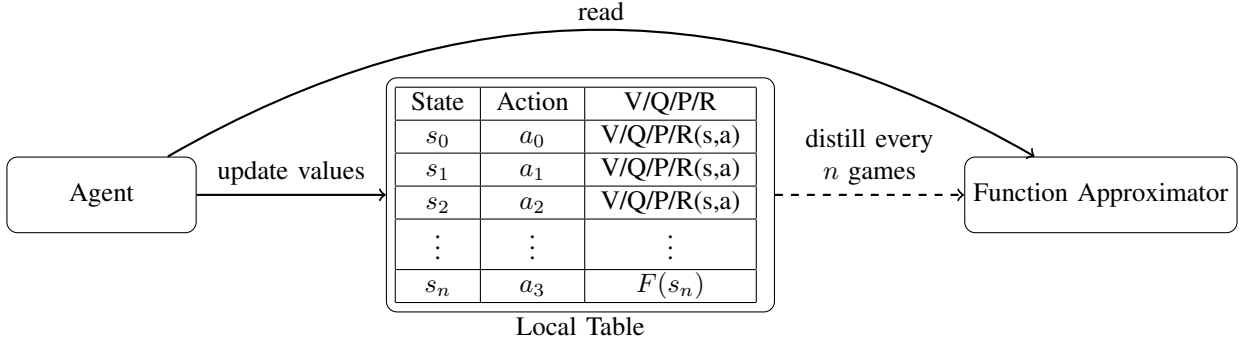
\begin{figure*}[h!]
\centering
\begin{tikzpicture}[node distance=25mm and 25mm]
\node[box] (agent) {Agent};
\node[box, right=of agent] (table) {
\begin{tabular}{|c|c|c|}
\hline
State & Action & V/Q/P/R \\ \hline
$s_0$ & $a_0$ & V/Q/P/R(s,a)\\ \hline
$s_1$ & $a_1$ & V/Q/P/R(s,a)\\ \hline
$s_2$ & $a_2$ & V/Q/P/R(s,a) \\ \hline
$\vdots$ & $\vdots$ & $\vdots$ \\ \hline
$s_n$ & $a_3$ & $F(s_n)$ \\ \hline
\end{tabular}
};
\node[box, right=of table] (dist) {Function Approximator};
\draw[arrow] (agent) -- (table) node[midway, above] {update values};
\draw[dasharrow] (table) -- (dist) node[midway, above, align=center] {distill every\\$n$ games};
\draw[arrow, bend left=30] (agent) to node[midway, above] {read} (dist);
\node[below=of table, yshift=73pt] {Local Table};
\end{tikzpicture}
\caption{The general LUGL architecture: an agent interacts with the environment (not shown), updates the appropriate local table, given the relevant algorithm, over states and actions, and periodically distils the table into a standard function approximator. }
  \label{fig:lugl}
\end{figure*}

Local Updates, Global Learning (LUGL) is a framework that builds reinforcement learning agents without the need for incremental learning. The core challenge it addresses is that non-incremental (batch) learners like gradient-boosted trees must be trained from scratch on a fixed dataset, whereas game-playing RL generates a continuously evolving, non-stationary distribution of states and targets. In standard settings, this mismatch causes severe distributional shift and instability. LUGL resolves this through two alternating phases that decouple data collection from model fitting.

At any point during training, a LUGL agent maintains two distinct representations of the learned quantity (Q-values, V-values, policies, or regrets depending on the variant). The first is a finite \emph{local table} (LT), which stores explicit entries for states or information sets encountered during the current self-play cycle. The second is a global \emph{function approximator} (FA), trained periodically from the contents of the local table. Only the local table is modified during self-play; the function approximator remains fixed until the next distillation step. In the algorithm descriptions that follow, the suffixes ``LT'' and ``FA'' are used to distinguish the local-table and function-approximator versions of a quantity, respectively.

\paragraph{Local Updates phase.}
 The agent plays self-play games using a policy derived from the current value estimates, and accumulates tabular updates into the \emph{local table}, exactly as in tabular RL.   Depending on the variant, the table stores Q-values, V-values, policy actions, or regret values. During this phase, only the local table is modified; the function approximator is not updated. What is being \emph{updated} is the tabular estimate of the relevant quantity (Q, V, policy, or regret) for each state (or information set) encountered during self-play. The update rules are standard (e.g., Q-learning, V-value updates, or CFR immediate regret computation), applied pointwise to table entries.

\paragraph{Global Learning phase.} After a fixed number of games (e.g., $10^4$ in our experiments), the local table is used as a supervised learning dataset to train the \emph{global function approximator} (e.g., LightGBM) that generalises the learned values to states not yet visited. The table entries serve as training samples: each row provides a state (or information set) representation $x$ as input features and the corresponding tabular value (Q, V, policy, or regret) as the target label. The function approximator $f(x)$ is trained to minimise the regression loss over this dataset. After training, the local table is discarded and rebuilt from scratch in the next cycle. What is being \emph{learned} is a global function $f \colon \mathcal{X} \to \mathbb{R}$ that maps state representations to value estimates, enabling the agent to evaluate states it has never encountered.

\paragraph{Stability mechanism.} The LUGL cycle stabilises non-incremental learners through two mechanisms. First, the local table acts as a buffer that decouples the data collection process from the model training process, analogous to experience replay in DQN but without maintaining a large persistent buffer. Second, the periodic distillation step ensures that the function approximator is always trained on a coherent snapshot of the current policy's performance, and the subsequent table reset prevents stale data from accumulating.

A schematic of the LUGL architecture can be seen in \figurename~\ref{fig:lugl}. The whole process starts with an initial function approximator and an empty local table, which is updated over $n$ games (in our case $10^4$). 
The policy is driven by the current hybrid value estimates, which use local-table values when available and otherwise fall back to the function approximator, while the training data used to update the function approximator comes from the local table.
Tables are capped to a maximum of $10^5$ sample rows, irrespective of the games played; beyond this size the memory footprint and LightGBM training time grow substantially with diminishing returns, as the current function approximator already covers the most frequently visited states.

\paragraph{State representation.} The function approximator requires fixed-length numerical feature vectors as input. For perfect-information games (AMGs), the state representation follows standard encodings: board configurations are represented as binary feature vectors (e.g., piece positions for Tic-tac-toe, Othello, and Hex; cell colours for Connect-4). For imperfect-information games, each information set is encoded as a tabular feature vector following the same convention used in the OpenSpiel DeepCFR implementation~\cite{lanctot2019openspiel}: one-hot encoded features for the player's private cards, the public board cards, and a positional encoding of the action history (e.g., the sequence of bets, calls, folds, and raises). Crucially, in the imperfect-information setting, we strip any information not available to the current player (e.g., the opponent's private cards), ensuring that multiple underlying game states that are indistinguishable to the player correctly map to the same information set and thus the same feature vector. This encoding approach is standard in the DeepCFR literature and is directly compatible with LightGBM's tabular input requirements.

The LUGL method can be adapted as a modification of various RL algorithms, using various function-approximation methods. In this paper we combine several standard RL methods with LightGBM and demonstrate a version of DeepCFR with LightGBM as well. The overall concept can be viewed as a modification of Fitted-Q iteration~\cite{antos2007fitted}

\subsection{Perfect information variations}

\subsubsection{\LUGLLightGBM{}}
In \LUGLLightGBM{} the local table and LightGBM model learn Q-values.  As with every other LUGL variant, the  Q-values are accumulated in the local table and periodically distilled into a function approximator. During self-play, updates are written to the local table while the function approximator provides value estimates for state-action pairs that are not yet represented in the table.

We denote the Local-Table and Function-Approximator versions of the Q-function by
$Q_\mathrm{LT}$ and $Q_\mathrm{FA}$, respectively, and define
\begin{equation}\widetilde{Q}(s,a):=\begin{cases}
Q_{\mathrm{LT}}(s,a), & \text{if } (s,a)\text{ exists in the local table}, \\
Q_{\mathrm{FA}}(s,a), & \text{otherwise},\end{cases}\label{eqn:QtablesHybrid}\end{equation}
as a hybrid of the two (which uses local-table estimates when available and otherwise falls back to the function approximator).

During exploration, the agent explores by choosing actions which are $\epsilon$-greedy on the hybrid $\widetilde{Q}$ function. For the LUGL method, we modify the standard update rule for Q-values \cite{Watkins_1992_QLearning}, such that whenever the agent transitions from state $s$ to state $s'$ after taking action $a$, the following update is made:
\begin{align} Q_\mathrm{LT}(s, a) \leftarrow \widetilde{Q}(s, a) + \alpha \left[ r + \gamma \max_{a'} \widetilde{Q}(s', a') - \widetilde{Q}(s, a) \right] \label{eqn:qupdate}\end{align}
where $\alpha>0$ is a learning rate,  $r$ is the immediate reward, and $\gamma$ is the discount factor.  Note that when a state-action pair does not yet exist in the local table, Equation~\eqref{eqn:qupdate} automatically initialises its value from $Q_\mathrm{FA}$ via the definition of $\widetilde{Q}$.

\subsubsection{\LUGLDLightGBM{}}
\LUGLDLightGBM{}  is a deterministic version of the previous algorithm (\LUGLLightGBM{}). Again the local table ($Q_\mathrm{LT}$) and LightGBM model ($Q_{\mathrm{FA}}$) store Q values, but now the agent performs deterministic Q-learning updates.

The agent updates its Q-values based on the maximum expected reward from the next state, using Equation \eqref{eqn:qupdate} with the learning rate $\alpha$ set to $1$. This deterministic approach simplifies the update process and can be effective in environments where the optimal policy is deterministic. When $\alpha=1$, the update rule for the Q-values \eqref{eqn:qupdate} reduces to:
\[ Q_\mathrm{LT}(s, a) \leftarrow \left[ r + \gamma \max_{a'} \widetilde{Q}(s', a') \right]. \]

The LightGBM model is used to approximate the $Q_\mathrm{FA}$ values.

\subsubsection{\LUGLPILightGBM{}}
In \LUGLPILightGBM{}, the local table and LightGBM model store both V-values and policy actions for each state. $V^{\pi}(s)$ denotes the value function under policy $\pi$. We denote the LT and FA versions of the value function and policy by $\{V^{\pi}_\mathrm{LT}, \pi_\mathrm{LT}\}$ and $\{V^{\pi}_\mathrm{FA},\pi_\mathrm{FA}\}$, respectively; and make definitions of $\widetilde{V}^{\pi}$ and $\widetilde{\pi}$ analogous to that of \eqref{eqn:QtablesHybrid}. The LightGBM agent performs approximate policy iteration, which involves iteratively updating the policy and value function. The agent first \textit{evaluates} the current policy using the LightGBM model to approximate the value function, via the update:

\[ V^{\pi}_\mathrm{LT}(s) \leftarrow \mathbb{E}_{\pi} \left[ r + \gamma \widetilde{V}^{\pi}(s') \mid s \right]. \]
Then it updates the policy via:\[ \pi_\mathrm{LT}(s) \leftarrow \arg\max_{a} \widetilde{V}^{\pi}(\oneStepLookaehadFunction(s,a)). \] where $\oneStepLookaehadFunction(s,a)$ is the one-step lookahead model for the game. As before, every $n$ games the local tables $\{V^{\pi}_\mathrm{LT}, \pi_\mathrm{LT}\}$ are distilled into LightGBM models and then reset.

\subsubsection{\LUGLVLightGBM{}}
In \LUGLVLightGBM{}, the local table ($V_\mathrm{LT}(s)$) and LightGBM model ($V_\mathrm{FA}(s)$) store V-values only, for each state.  The agent updates its value function based on the expected reward from the next state, by the rule:
\[ V_\mathrm{LT}(s) \leftarrow \widetilde{V}(s) + \alpha \left[ r + \gamma \widetilde{V}(s') - \widetilde{V}(s) \right], \]
where $s$ is the current state, $r$ is the immediate reward, $s'$ is the next state, $\alpha$ is the learning rate, $\gamma$ is the discount factor, and $\widetilde{V}$ is defined analogously to \eqref{eqn:QtablesHybrid}. The policy followed is typically a policy that considers the exploration-exploitation trade-off, such as an $\epsilon$-greedy one-step-lookahead policy based on the successor-state values predicted by $\widetilde{V}$. This approach simplifies the update process, and can be effective in environments where the optimal policy is deterministic.

\begin{figure*}[t!]
  \centering
  \begin{subfigure}[b]{0.45\linewidth}
    \includegraphics[width=\linewidth]{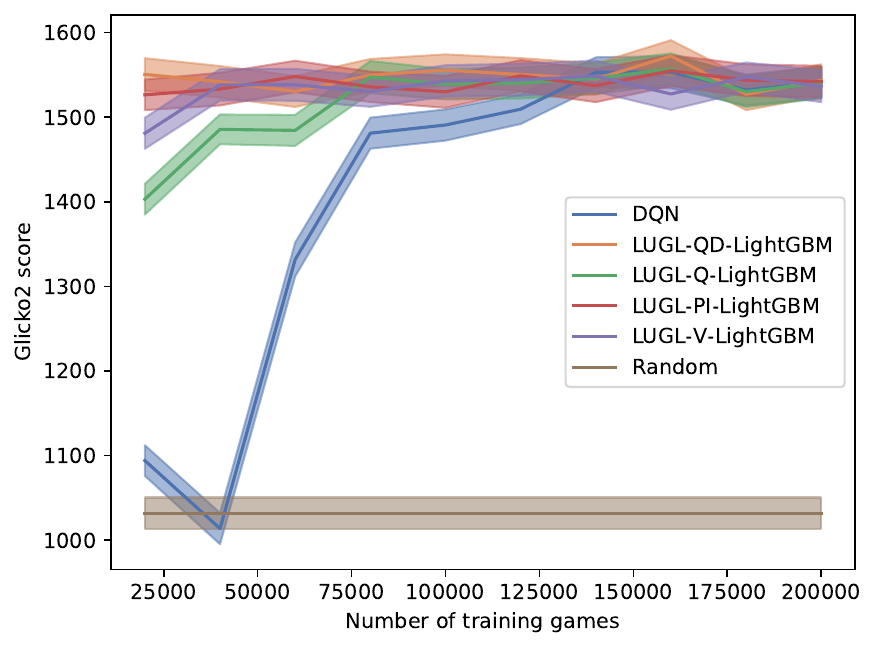}
    \caption{Tic-tac-toe}
    \label{fig:tic}
  \end{subfigure}
  \hspace{\fill}
  \begin{subfigure}[b]{0.45\linewidth}
    \includegraphics[width=\linewidth]{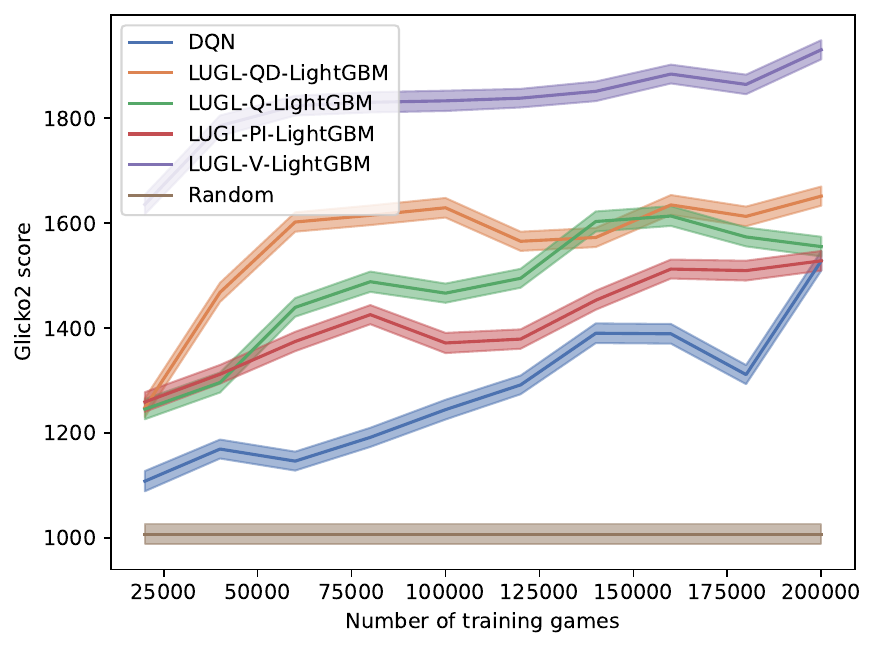}
    \caption{Connect Four}
    \label{fig:cf}
  \end{subfigure}
  \vskip\baselineskip
  \begin{subfigure}[b]{0.45\linewidth}
    \includegraphics[width=\linewidth]{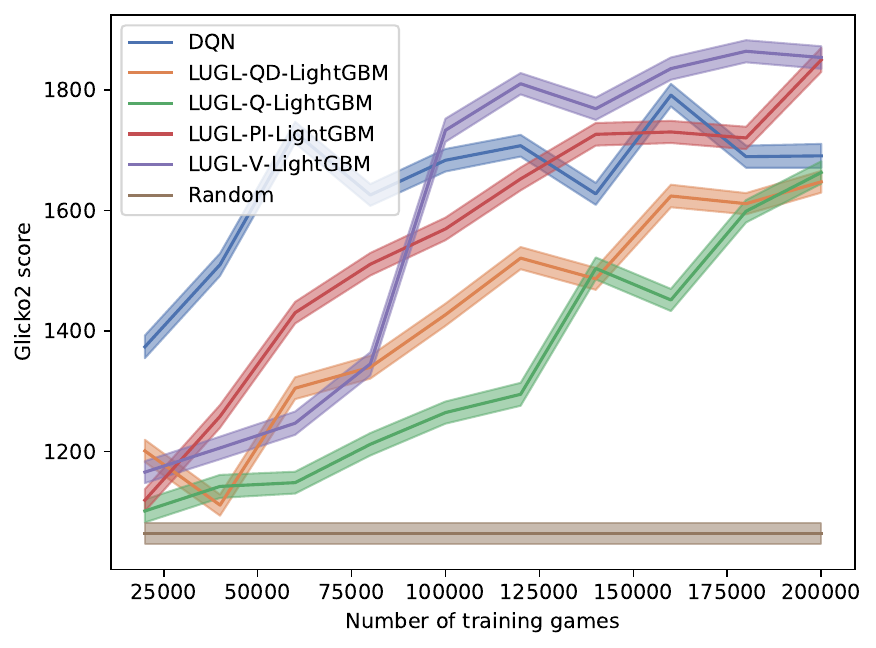}
    \caption{Othello}
    \label{fig:othello}
  \end{subfigure}
  \hspace{\fill}
  \begin{subfigure}[b]{0.45\linewidth}
    \includegraphics[width=\linewidth]{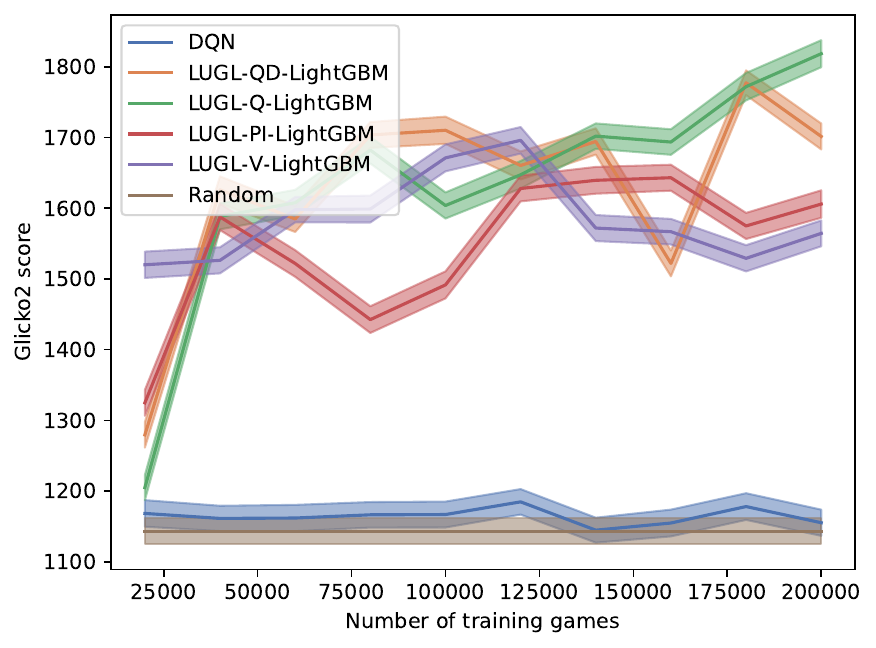}
    \caption{Hex}
    \label{fig:hex}
  \end{subfigure}
  \caption{Glicko-2 scores for different agents. Notice the short term gains of all LUGL variants vs DQN. }
  \label{fig:allperfect}
\end{figure*}

\subsection{Imperfect information variations}
\subsubsection{\LUGLDeepCFRLightGBM{}}
In \LUGLDeepCFRLightGBM{}, the local table and LightGBM model store Regret values.  This algorithm is based on DeepCFR, but with a key modification: we replace the neural networks with LightGBM models. Since the original DeepCFR approach trains its neural networks from scratch at every iteration, switching to LightGBM does not introduce additional complications, and in fact one could argue that DeepCFR is closer in spirit to our LUGL framework than it is to DQN.

The training procedure follows the standard DeepCFR iterative process adapted to the LUGL framework. During the local updates phase, trajectory sampling generates game trajectories using an external sampling scheme: all possible actions are explored for the currently active player, while only a single action is sampled for all other players and chance events. For each sampled trajectory, immediate regret values are computed and stored in the local table, indexed by information set. In the imperfect information setting, the state is replaced by the information set defined in Section~\ref{Sec:Background}. An information set groups together all game states that a player cannot distinguish given their observations; thus, the input to the function approximator is the observable history (e.g., in poker, one's own hole cards, the public board, and the sequence of actions), with the opponent's private cards excluded. This means we take the collected states and remove any information unavailable to the player (e.g., the opponent's private cards), which causes multiple underlying game states to map to the same information set.

After a fixed number of traversals, the global learning phase trains a LightGBM model from the accumulated regret samples in the local table, using the same feature encoding described in Section~\ref{Sec:Methodology}: one-hot encoded private cards, public board cards, and action history. The trained model approximates cumulative regret, from which the current policy is derived via regret-matching. The table is then reset and the cycle repeats. This substitution allows us to take advantage of the efficiency and stability of gradient-boosted decision trees, while maintaining the overall structure of the DeepCFR algorithm.

\begin{figure*}[t!]
  \centering
  \begin{subfigure}[b]{0.45\linewidth}
    \includegraphics[width=\linewidth]{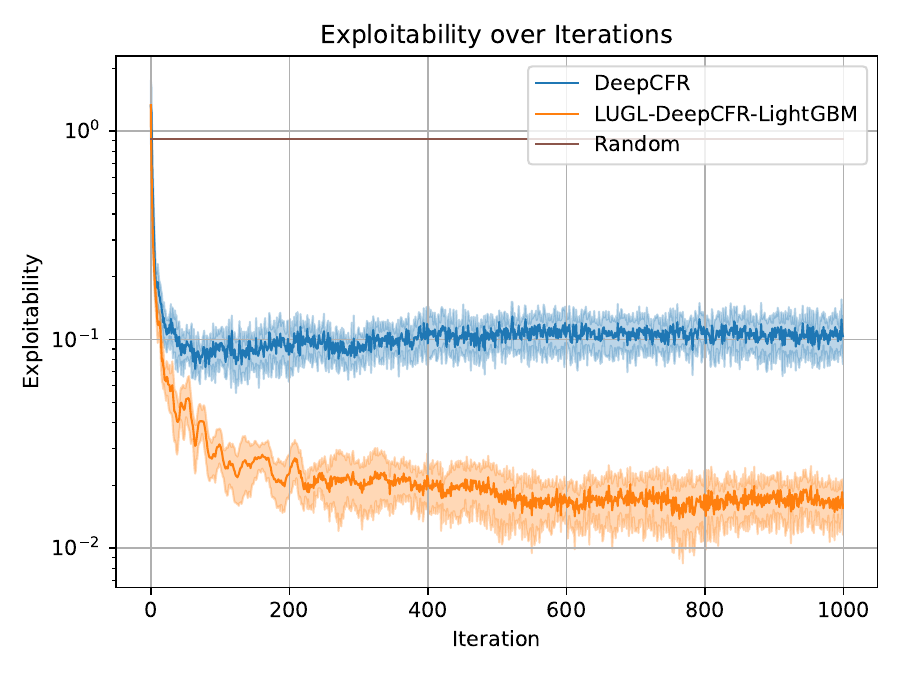}
    \caption{Exploitability in Kuhn poker.}
    \label{fig:kuhn}
  \end{subfigure}
  \begin{subfigure}[b]{0.45\linewidth}
    \includegraphics[width=\linewidth]{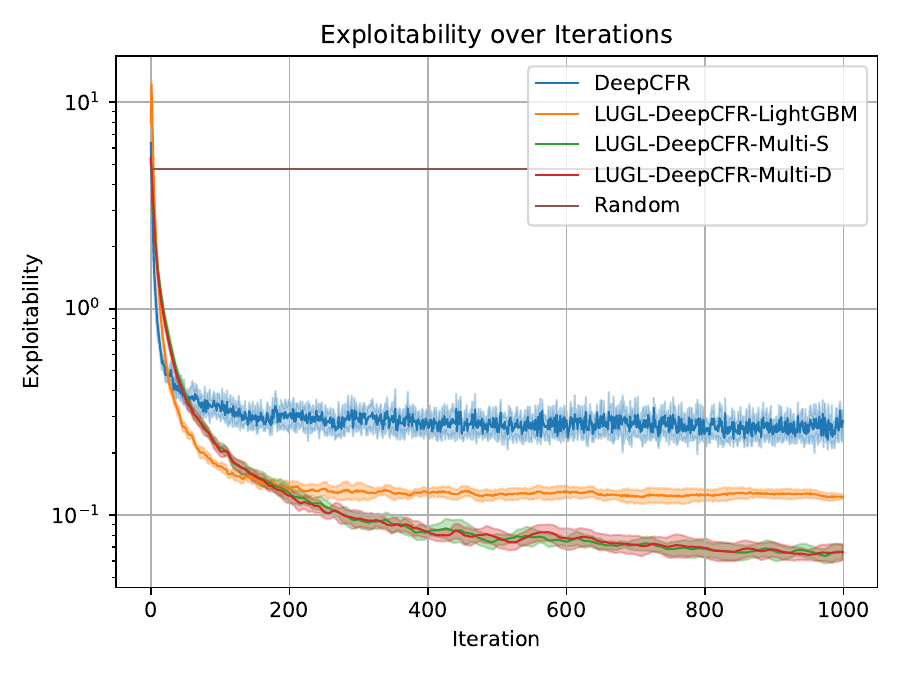}
    \caption{Exploitability in Leduc poker.}
    \label{fig:leduc}
  \end{subfigure}
  \begin{subfigure}[b]{0.45\linewidth}
    \includegraphics[width=\linewidth]{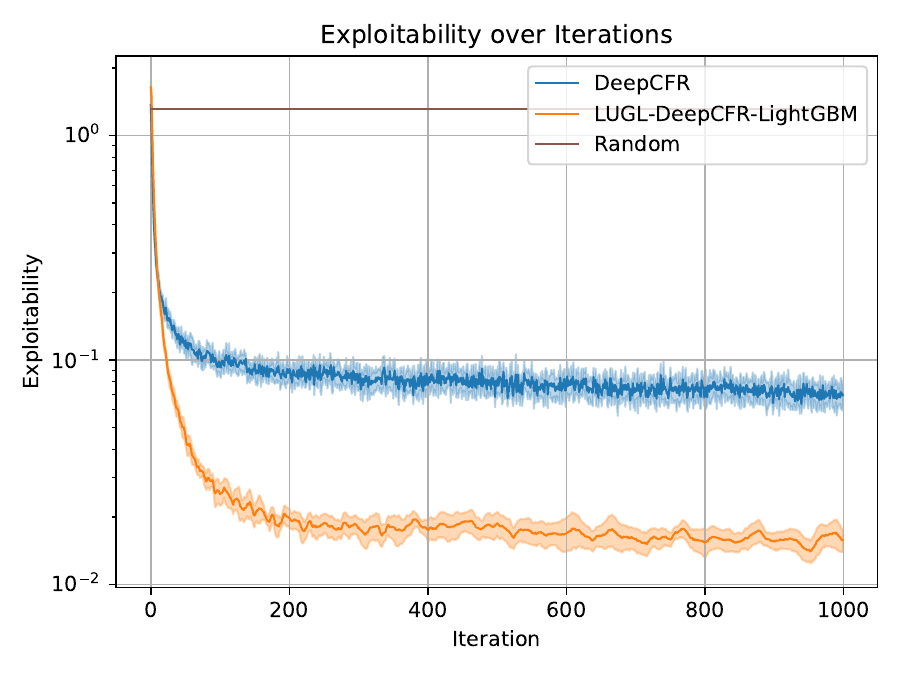}
    \caption{Exploitability in Liar's Dice.}
    \label{fig:liars_dice}
  \end{subfigure}
  \begin{subfigure}[b]{0.45\linewidth}
    \includegraphics[width=\linewidth]{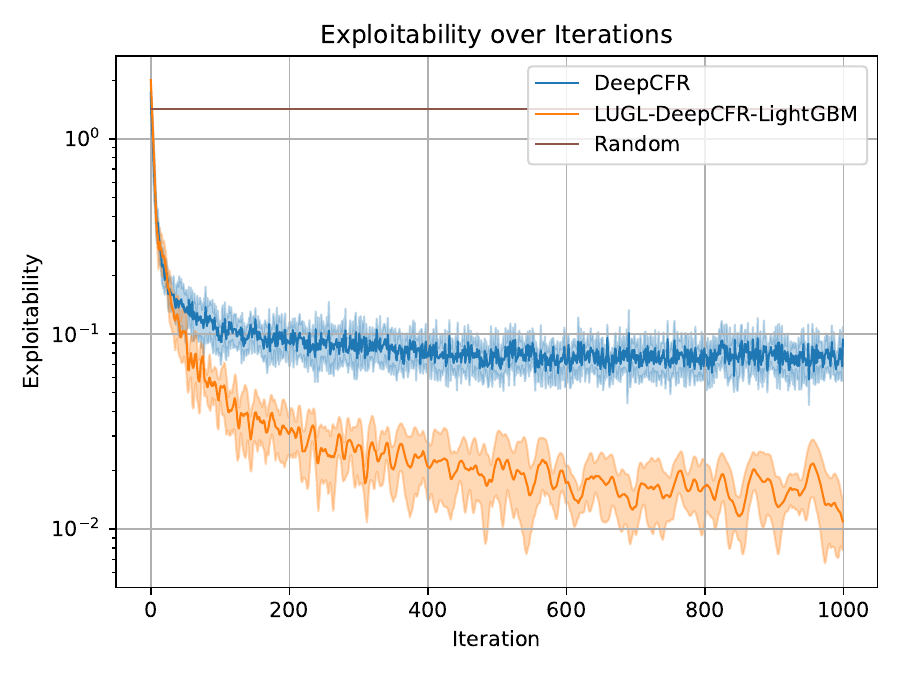}
    \caption{Exploitability in imperfect information Goofspiel 4.}
    \label{fig:goofspiel}
  \end{subfigure}
  \caption{Exploitability comparison of DeepCFR and LUGLDeepCFR variants on different imperfect information games. }
  \label{fig:imperfect_first_experiments_subs}
\end{figure*}

\begin{figure*}[t!]
  \centering
  \begin{subfigure}[b]{0.45\linewidth}
    \includegraphics[width=\linewidth]{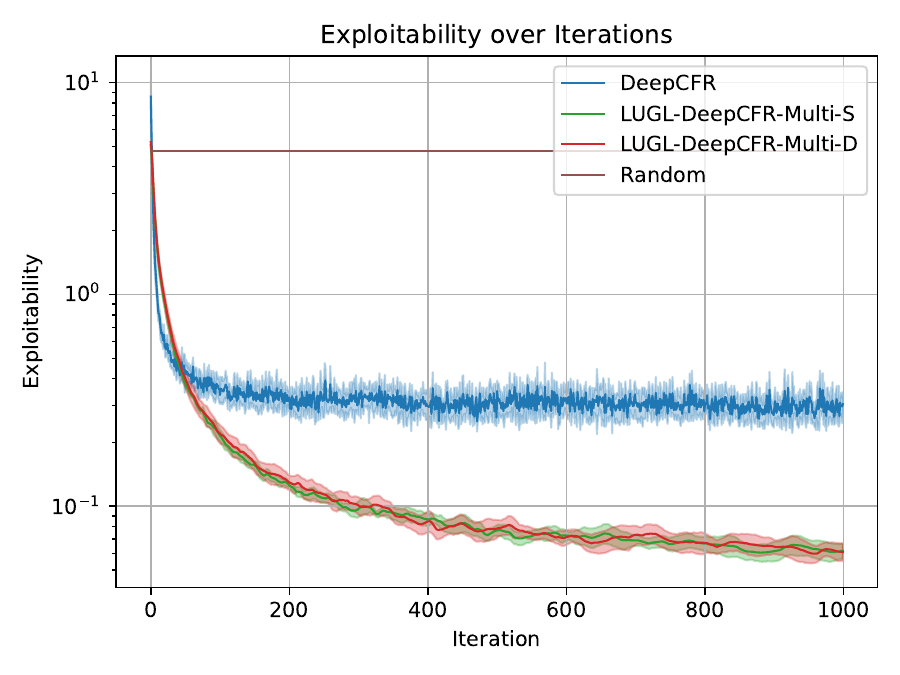}
    \caption{Jack and King left out.}
    \label{fig:jack_and_king}
  \end{subfigure}
  \begin{subfigure}[b]{0.45\linewidth}
    \includegraphics[width=\linewidth]{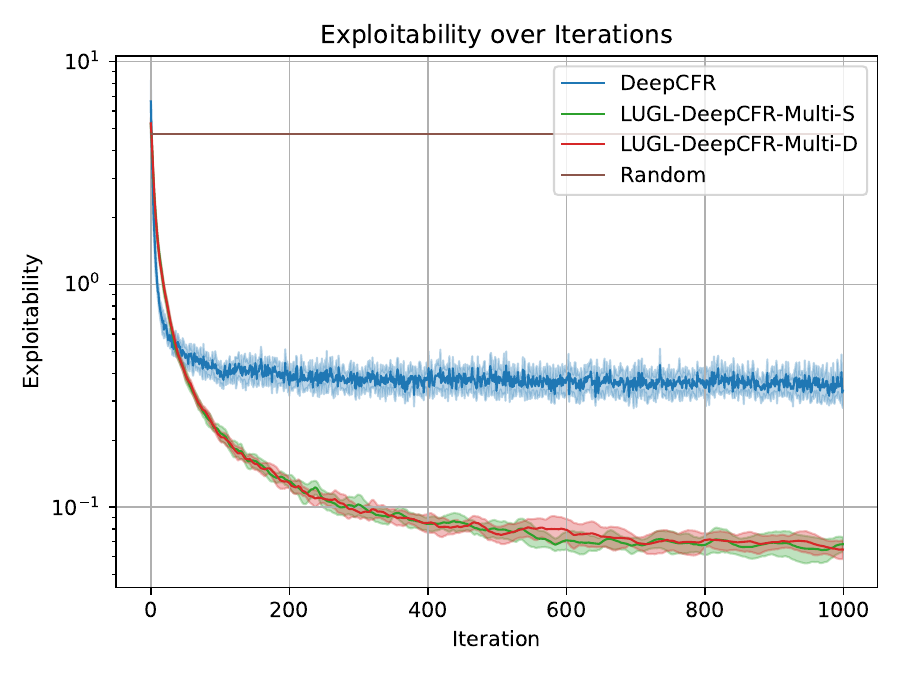}
    \caption{Jack and Queen left out.}
    \label{fig:jack_and_queen}
  \end{subfigure}
  \begin{subfigure}[b]{0.45\linewidth}
    \includegraphics[width=\linewidth]{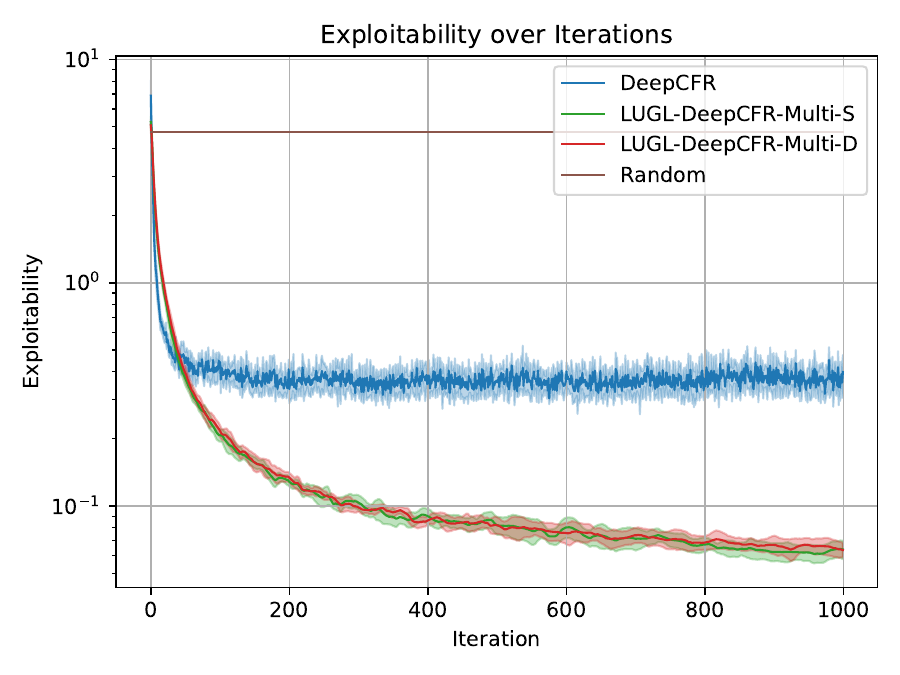}
    \caption{Queen and King left out.}
    \label{fig:queen_and_king}
  \end{subfigure}
  \begin{subfigure}[b]{0.45\linewidth}
    \includegraphics[width=\linewidth]{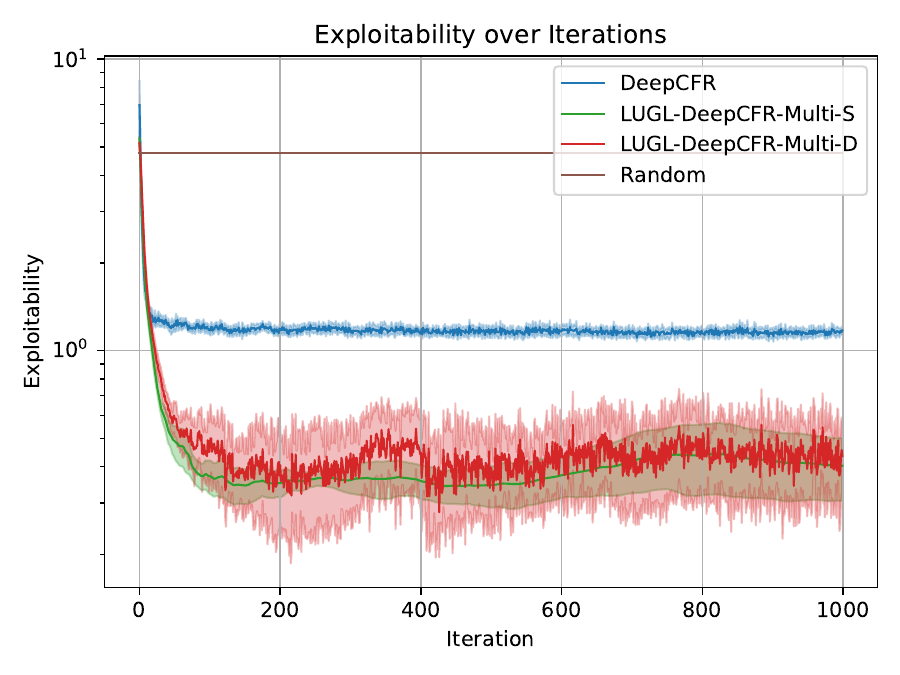}
    \caption{Queens left out.}
    \label{fig:queens}
  \end{subfigure}
  \caption{Generalisation capability of DeepCFR and LUGLDeepCFRMulti* in Leduc Poker when leaving out different states in the game when training.}
  \label{fig:allsubs}
\end{figure*}

\subsubsection{\LUGLDeepCFRMultiSD{}}
The \LUGLDeepCFRMultiSD{} variants extend DeepCFR further by replacing a single global approximator with a collection of smaller, specialized models. Each model corresponds to a specific betting sequence in Leduc Poker. During training, data collected in the reservoir buffer is grouped by betting sequence. Each group is then used to train a separate approximator dedicated to that particular sequence. This approach enables more precise modeling of different parts of the game. The two variants differ only in the type of approximator used, and notably, we do not use LightGBM. Because the task is split across many specialized models, each handles a smaller subset of the data, allowing the use of simpler approximators. \textbf{\LUGLDeepCFRMultiS{}} uses polynomial function approximators with splines and \textbf{\LUGLDeepCFRMultiD{}} uses decision trees as approximators. By tailoring models to specific betting sequences, these variants aim to improve both accuracy and generalization compared to a single, global approximator.

\section{Experiments}

We used OpenSpiel~\cite{lanctot2019openspiel} for all experiments. The goal of the experimental setup is to show the strengths and any potential limitations of the current approach.

\subsection{Perfect information}
We ran all experiments comparing the performance of our variants vs a random player and standard deep Q-learning as a baseline. Agent snapshots taken at fixed intervals play in an all-play-all round-robin tournament; Glicko-2 ratings are computed from the resulting match outcomes. LUGL and LightGBM hyperparameters are $n\_distillation = 10^4$, 
$n\_measure\_games = 64$  and $n\_trees$ = 2000. Looking at different game performance in Figure~\ref{fig:allperfect}, one can see how rapidly LUGL picks up vs DQN (as expected, effectively LightGBM trains each iteration to convergence). Certain variations of the algorithm perform better in different games, with the version approximating the V-value function performing better in two different cases. This might be due to the fact that the V-value is easier to approximate, though a V-value setup might make the algorithm harder to use in non-deterministic settings.  Note that in Hex, \LUGLVLightGBM{} begins to plateau after approximately 125K games while DQN continues to improve; this is likely due to the interaction between the table size cap and the increased depth of lookahead required as the game progresses. Overall, we tend to get ``reasonable'' quality players very early in training.

\subsection{Imperfect Information}
We ran experiments on a cluster with Intel Xeon Scalable Gold 6146 processors, using 4 threads and 64 GB of memory. Hyperparameters for all experiments are reported in Table~\ref{tab:ii_hyperparams}. We adopted the hyperparameters from \cite{walton2021multi} but significantly reduced them for Kuhn Poker, as it is a much smaller game. Since DeepCFR retrains the network using $N$ steps with a batch size of $K$, we sample $N \times K$ data points for LUGLDeepCFR* to ensure a fair comparison between experiments.


\subsubsection{Performance Comparison Across Games}

We tested DeepCFR against our proposed variant (\LUGLDeepCFRLightGBM{}) across four imperfect-information games: Kuhn Poker, Leduc Poker, Liar's Dice, and IIGoofspiel(4), and we show the results in Figure~\ref{fig:imperfect_first_experiments_subs}. The results demonstrated that replacing neural networks with LightGBM in the \LUGLDeepCFRLightGBM{} variant consistently outperformed the original DeepCFR algorithm across all tested games.

LightGBM allowed for much higher exploitability reduction, confirming that LightGBM is a strong alternative to deep neural networks for the regret and policy approximation tasks within the DeepCFR framework.

For the \LUGLDeepCFRMultiS{} and \LUGLDeepCFRMultiD{} variants, we only ran experiments on Leduc Poker, as this game provides a manageable yet non-trivial testbed with a well-defined set of betting sequences. These multi-variants significantly outperformed even the LightGBM version, demonstrating the benefits of splitting the approximation task into smaller, specialized models for each betting sequence. The polynomial spline-based MultiS and the decision tree-based MultiD both showed very similar strong performance as shown in Figure~\ref{fig:leduc}.

\subsubsection{Generalization Experiments in Leduc Poker}

We also conducted generalization experiments to test how well the algorithms perform on previously unseen states. In these experiments, we deliberately excluded specific card combinations from the training data so that the algorithms never encountered them during training. Specifically, we withheld all game states involving:
\begin{itemize}
    \item Jack and King in Figure~\ref{fig:jack_and_king},
    \item Jack and Queen in Figure~\ref{fig:jack_and_queen},
    \item Queen and Jack in Figure~\ref{fig:queen_and_king},
    \item Two Queens in Figure~\ref{fig:queens}.

\end{itemize}
This setup forced the algorithms to generalise their learned policies to novel situations.

Across all withheld state scenarios, both DeepCFR and the Multi variants performed worse than in the standard setting, confirming the difficulty of generalisation in imperfect-information games. However, the Multi variants (both MultiS and MultiD) still significantly outperformed DeepCFR, showing that specialised approximators per betting sequence can generalize more effectively than a single global approximator.

An interesting and unexpected result occurred in the case where the withheld states involved two Queens. Here, DeepCFR performed extremely poorly, failing to learn a coherent strategy, while the Multi variants exhibited very high variance between runs. This suggests that these withheld states posed a particularly difficult generalisation challenge, possibly because of their strategic importance.

These experiments were designed to test whether the use of multiple specialised approximators improves generalisation. While the performance drop confirmed the difficulty of generalising to unseen game states, the superior performance of the Multi variants indicates that the decomposition of the approximation problem is a promising direction for handling large, complex imperfect-information games.

\begin{table}[h!]
\centering
\normalsize
\caption{Hyperparameters for OpenSpiel DeepCFR. The first column lists values used for Kuhn Poker, and the second column lists values for Leduc Poker, Liar's Dice, and IIGoofspiel 4. These also apply to LUGLDeepCFR* where relevant.}
\label{tab:ii_hyperparams}
\begin{tabular}{lll}
\toprule
\textbf{Parameter} & \textbf{Kuhn} & \textbf{Other} \\ \midrule
num\_traversals & 50 & 1 500 \\
batch\_size\_advantage & 64 & 2 048 \\
batch\_size\_strategy & 64 & 2 048 \\
num\_hidden & 32 & 64 \\
num\_layers & 2 & 3 \\
reinitialize\_advantage\_networks & \multicolumn{2}{c}{True} \\
learning\_rate & \multicolumn{2}{c}{1e-3} \\
memory\_capacity & \multicolumn{2}{c}{1 000 000} \\
policy\_network\_train\_steps & 500 & 5000 \\
advantage\_network\_train\_steps & 100 & 750 \\ \bottomrule
\end{tabular}
\end{table}

\subsubsection{Flop5 Hold'em}
For the larger experiments, we used cluster nodes equipped with AMD EPYC 7543 CPUs. All experiments were executed using 64 CPU threads, 256\,GB of RAM, and an NVIDIA Tesla A100 GPU for neural network training.

As a baseline, we used Single Deep CFR (SD-CFR) \cite{steinberger2019single, steinberger2019pokerrl} instead of Deep CFR. SD-CFR avoids training a separate average policy network and instead stores approximators from all previous iterations, training on them. The final policy is a weighted average of the decisions of the stored approximators, which in CFR would represent the ``current policies'' from previous iterations.

We performed five independent runs of SD-CFR and ten runs of \LUGLDeepCFRLightGBM{}. Initially, five of the \LUGLDeepCFRLightGBM{} runs were configured to use approximately the same amount of training data per iteration as the neural network baseline (4\,096\,000 samples for the neural network and 4\,000\,000 for \LUGLDeepCFRLightGBM{}). Under this configuration, \LUGLDeepCFRLightGBM{} training was approximately 15 times slower. To improve runtime, we reduced the maximum training data for \LUGLDeepCFRLightGBM{} to 1\,000\,000 samples and adjusted additional hyperparameters (details are provided in the Appendix). This reduced the runtime overhead to approximately 1.5$\times$ slower than the neural network baseline. Considering that the implementation is optimized for neural networks and \LUGLDeepCFRLightGBM{} was used with minimal modifications, this slowdown is acceptable.

\begin{figure}
    \centering
    \includegraphics[width=\linewidth]{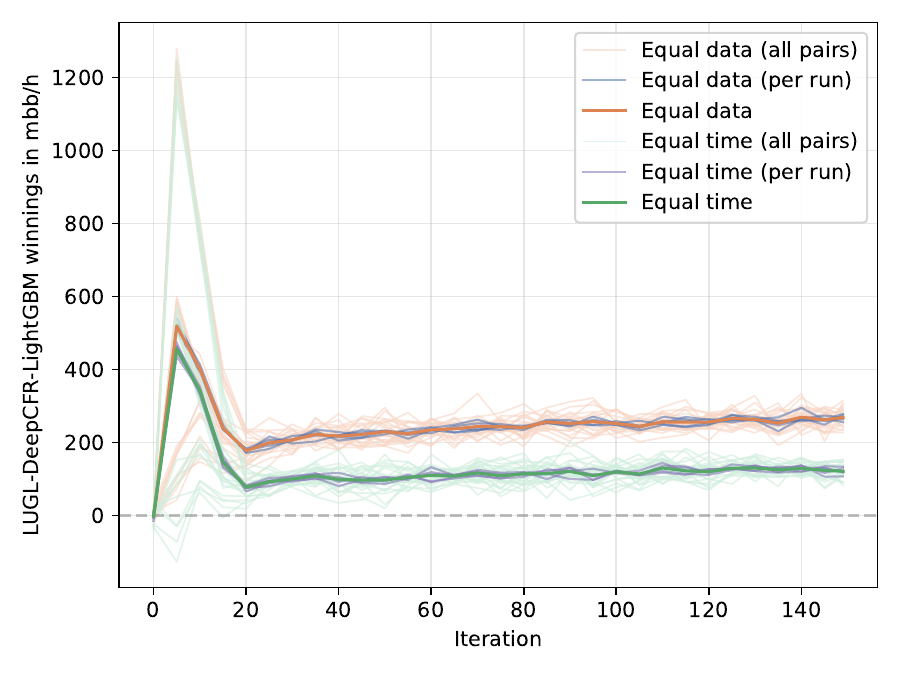}
    \caption{Winnings of the LUGL-DeepCFR-LightGBM vs DeepCFR. Both algorithms use the SD-CFR variant.}
    \label{fig:flop5holdem}
\end{figure}

To evaluate performance, agents from each run were matched against each other every five iterations, resulting in 25 head-to-head matches per evaluation point. Each match consisted of two million hands. Figure~\ref{fig:flop5holdem} reports the individual match results, the per-run average for \LUGLDeepCFRLightGBM{}, and the overall average.

The results show an initial spike in the average win rate of \LUGLDeepCFRLightGBM{}. However, the individual trajectories indicate that this value is noisy, and the similarity of the per-run averages suggests that the spike is likely caused by some SD-CFR runs having weaker early iterations. After approximately iteration 20, the results stabilize. From this point onward, \LUGLDeepCFRLightGBM{} shows a small but consistent improvement in win rate, consistent with the smaller-scale experiments, where DeepCFR appears to plateau while \LUGLDeepCFRLightGBM{} continues to improve.

We additionally evaluated smaller and larger variants of the neural network architecture to determine whether SD-CFR performance could be improved by adjusting model capacity. Both alternatives resulted in reduced performance, indicating that the original architecture is well-suited to the task. We provide details in the appendix.

\begin{table}[h!]
\centering
\normalsize
\caption{Hyperparameters for Single DeepCFR. Max buffer size is per worker, so with our setup, the total buffer size is 124 million.}
\label{tab:flop5_hyperparams}
\begin{tabular}{lll}
\toprule
\textbf{Parameter} & \textbf{Value} \\ \midrule
n\_learner\_actors & 62 \\
n\_traversals\_per\_iter & 30 000 \\
n\_batches\_adv\_training & 2 000 \\
batch\_size & 2 048 \\
max\_buffer\_size & 2 000 000 \\
lr & 0.001 \\
n\_units\_final & 64 \\
use\_pre\_layers & True \\
card\_block\_units & 192 \\
other\_units & 64 \\ \bottomrule
\end{tabular}
\end{table}

\section{Conclusion}
\label{sec:conclusion}
This paper introduced LUGL, a framework enabling non-incremental learners, specifically gradient-boosted trees, to achieve competitive performance in reinforcement learning environments. The stability of LUGL in iterative learning tasks arises from two mechanisms: (i) the local table decouples data collection from model training, acting as a buffer that prevents distributional shift between successive training cycles, and (ii) the periodic distillation step ensures that the function approximator is always trained on a coherent snapshot of the current policy's performance, with the table reset preventing the table from growing indefinitely. Empirically, this yields smooth, monotonic improvement in Glicko-2 ratings for perfect-information games,  consistent exploitability reduction for imperfect-information games and also greater robustness and generalization capabilities. 

Our results show that LUGL frequently outperforms standard neural-network-based methods in terms of solution quality. In perfect-information games such as Tic-tac-toe, Connect-4, Othello, and Hex, LUGL variants consistently achieved high win rates in head-to-head comparisons, demonstrating rapid convergence compared to DQN baselines. In the domain of imperfect-information games---including Kuhn Poker, Leduc Holdem, Liar's Dice, and Goofspiel---LUGL-CFR consistently reached significantly lower exploitability levels than DeepCFR. Notably, in the large-scale Flop5 Hold'em domain, our experiments showed a substantial performance advantage, with LUGL-CFR outperforming the Single Deep CFR (SD-CFR) baseline by 100 mbb/h in head-to-head winnings. These findings challenge the assumption that incremental learning is required for success in complex games and establish gradient-boosted trees as a powerful, non-parametric alternative for RL tasks with structured or tabular data.

It is worth noting that the approach pioneered here is still within the statistical realm. For strong(er) generalisation capacity, one might need to combine it with methods inspired by old-school game solutions~\cite{allis1988knowledge}, where one would extract global proofs from trajectories, and thus generate  exceptional features that can be used in the game irrespective of position. This would point to methods that can extract global rules like inductive logic programming~\cite{muggleton1991inductive}. 
\section*{Acknowledgments}

This article was supported by COST Action CA22145 - Computational Techniques for Tabletop Games Heritage (GameTable), supported by COST (European Cooperation in Science and Technology) and by the Czech Science Foundation (grant no. GA25-18353S). Computational resources were supplied by the OP VVV funded project CZ.02.1.01/0.0/0.0/16\_019/0000765 ``Research Center for Informatics''. We also acknowledge the support of Engineering and Physical Sciences Research Council (EPSRC) grant ''UKRI1076 Element-compositor methods for out-of-distribution machine learning''.  ChatGPT~\cite{chatgpt} was used to help edit the background section.

\ifCLASSOPTIONcaptionsoff
  \newpage
\fi



\bibliographystyle{IEEEtran}
\bibliography{refs,Dennis-Soemers-Bib}







\begin{appendices}
\section{LightGBM Hyperparameters}
The table summarizes the LightGBM hyperparameter configurations used across different experimental domains. The \textit{Small} setting represents the standard configuration applied to all games except Flop5 Holdem. The \textit{Time} setting was tuned to produce runs with a similar computation time as SD-CFR, whereas the \textit{Data} setting used an equivalent amount of training data as SD-CFR, resulting in a significantly longer runtime.
\begin{table}[h!]
\centering
\normalsize
\caption{Hyperparameters for LightGBM in the different domains. Small is the setting used for all the games except Flop5 Holdem. Time is the setting used to create runs with a similar time as the SD-CFR, and the Data is the setting that used the same amount of data as the SD-CFR but was 15 times slower.}
\label{tab:lgbm_hyperparams}
\begin{tabular}{lccc}
\toprule
\textbf{Parameter} & \textbf{Small} & \textbf{Time} & \textbf{Data} \\
\midrule
num\_boost\_round      & 200       & 100        & 300 \\
learning\_rate         & 0.1       & 0.1        & 0.1 \\
num\_leaves            & 64        & 64         & 96 \\
max\_depth             & 7         & 6          & 7 \\
min\_data\_in\_leaf    & 20        & 20         & 20 \\
bagging\_fraction      & 0.8       & 0.8        & 0.8 \\
bagging\_freq          & 1         & 1          & 1 \\
feature\_fraction      & 0.8       & 0.8        & 0.8 \\
lambda\_l1             & 0.1       & 0.1        & 0.1 \\
lambda\_l2             & 0.1       & 0.1        & 0.1 \\
\bottomrule
\end{tabular}
\end{table}

\section{Results with Different Neural Network Capacities}

\begin{figure}[t]
    \centering
    \includegraphics[width=\linewidth]{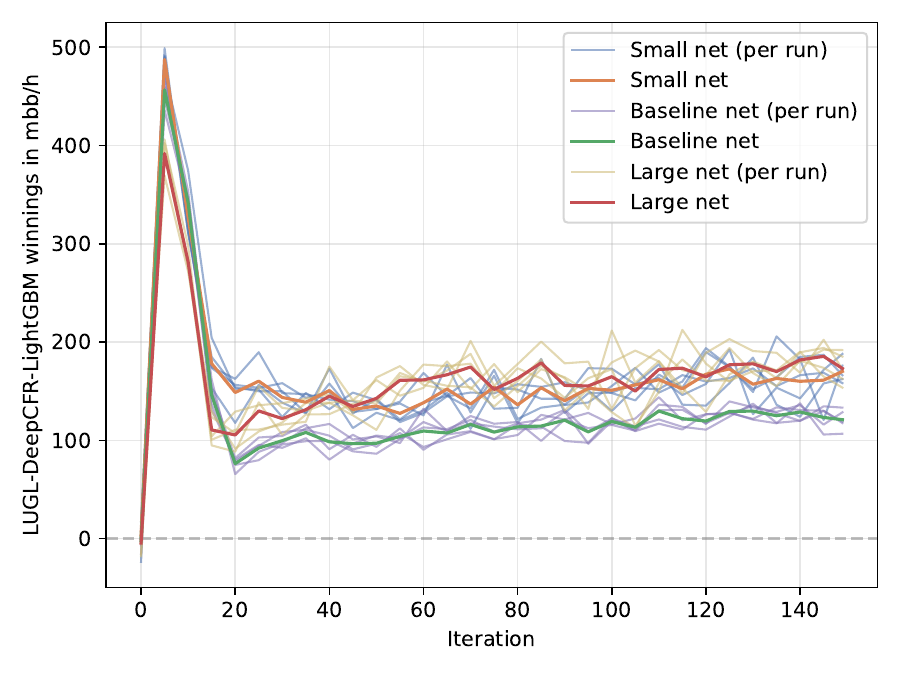}
    \caption{Study of the impact of network capacity on the performance of the SD-CFR -- lower values are better.}
    \label{fig:net_sizes}
\end{figure}

The neural network used in SD-CFR follows the architecture proposed in the Deep CFR paper \cite{brown2019deep}. To evaluate the effect of model capacity, we trained both smaller and larger variants of this architecture.
The smaller network uses 96 card block units and 32 units for all the other layers. The baseline network uses 192 card block units and 64 units for all the other layers, while the larger network uses 384 card block units and 128 units in the corresponding layers.
These configurations result in approximately 30\,000 parameters for the small network, 111\,000 parameters for the baseline network, and 418\,000 parameters for the large network.
For comparison, we performed three independent training runs for each configuration. The results show that both the smaller and larger networks perform worse than the baseline model, as illustrated in Figure~\ref{fig:net_sizes}. 

\end{appendices}

\end{document}